\documentclass{article} 
\usepackage{PRIMEarxiv}

\usepackage{amsmath,amsfonts,bm}

\def\eqref#1{equation~\ref{#1}}

\def\1{\bm{1}}

\DeclareMathAlphabet{\mathsfit}{\encodingdefault}{\sfdefault}{m}{sl}
\SetMathAlphabet{\mathsfit}{bold}{\encodingdefault}{\sfdefault}{bx}{n}

\usepackage{hyperref}
\usepackage{url}
\usepackage{graphicx}
\usepackage{booktabs}
\usepackage{caption}
\usepackage{xspace}
\usepackage{dsfont}
\usepackage{amssymb}

\usepackage{bbm}
\usepackage{multirow}
\usepackage{subcaption}
\usepackage{makecell}
\usepackage{float}

\usepackage{xcolor}
\hypersetup{
	colorlinks,
	linkcolor={red!50!black},
	citecolor={blue!50!black},
	urlcolor={blue!80!black}
}

\makeatletter
\DeclareRobustCommand\onedot{\futurelet\@let@token\@onedot}
\def\@onedot{\ifx\@let@token.\else.\null\fi\xspace}

\def\eg{{e.g}\onedot} 
\def\ie{{i.e}\onedot}

\makeatother

\usepackage{natbib}
\title{Same accuracy, twice as fast: continuous training surpasses retraining from scratch}

\author{Eli Verwimp$^1$ \\
KU Leuven \\
\texttt{eli.verwimp@kuleuven.be} \\
\And
Guy Hacohen$^1$\\
KU Leuven \\
\texttt{guy.hacohen@kuleuven.be} \\
\AND
Tinne Tuytelaars \\
KU Leuven \\
\texttt{tinne.tuytelaars@kuleuven.be} \\
}

\begin{document}

\maketitle
\footnotetext[1]{Both authors contributed equally.}

\begin{abstract}
Continual learning aims to enable models to adapt to new datasets without losing performance on previously learned data, often assuming that prior data is no longer available. However, in many practical scenarios, both old and new data are accessible. In such cases, good performance on both datasets is typically achieved by abandoning the model trained on the previous data and re-training a new model from scratch on both datasets. This training from scratch is computationally expensive. In contrast, methods that leverage the previously trained model and old data are worthy of investigation, as they could significantly reduce computational costs. Our evaluation framework quantifies the computational savings of such methods while maintaining or exceeding the performance of training from scratch. We identify key optimization aspects -- initialization, regularization, data selection, and hyper-parameters -- that can each contribute to reducing computational costs. For each aspect, we propose effective first-step methods that already yield substantial computational savings. By combining these methods, we achieve up to 2.7x reductions in computation time across various computer vision tasks, highlighting the potential for further advancements in this area.
\end{abstract}

\section{Introduction}
In machine learning applications, the available data tends to expand and evolve over time. This often requires updating a model that was trained on a large dataset (`\emph{old data}') to be further trained and adapted to also perform well on a new dataset (`\emph{new data}'). Such new datasets often include data from different data generating distributions, which may entail additional classes, new domains or corner cases. In practical scenarios, a common approach is to retrain the model from scratch using both the old and new dataset. Yet when model and dataset sizes increase, retraining from scratch becomes computationally expensive. Reducing these costs requires more efficient methods to train models continuously, starting from a model trained on a part of the full dataset.

A large part of continual learning research approached this problem in a resource-constrained setting, aiming to reach the highest possible performance on both old and new data under some constraints~\citep{verwimpcontinual}. A common constraint is to limit the amount of memory usage, which restricts how much old data can be stored. Such constraints can lead to suboptimal performance, in part due to catastrophic forgetting~\citep{french1999catastrophic}. In \eg industry applications, it is often undesirable to sacrifice performance and retraining from a randomly initialized model (`\emph{from scratch}') is preferred over using older models~\citep{Huyen_2022}. This is a wasteful approach; there is a model available that performs well on the old data, so why not use it? In practice, it has been shown to be difficult to continue to train previously trained models, even without storage constraints on past examples~\citep{ash2020warm}.

In this paper, we aim to reduce the cost of training a model on new data, when it has already been trained on some old data. The cost of training this old model is treated as a sunk cost~\citep{kahneman1972subjective}, its training happened in the past and the price for this training has already been paid. In contrast, future costs for training on new data can be reduced or mitigated, and in this paper, we show that using old models can be an effective way of doing so. In many computational problems, the total cost consists of both memory and computational aspects. However, for the size of modern networks, the computational costs tend to outweigh the memory costs. For example, the hard disks to store ImageNet21k are about 50 times cheaper than training a large vision transformer on the same data once (see Appendix~\ref{sec:cost_estimation} for details of this estimate).

\begin{figure}
	\centering
	\includegraphics[width=0.9\linewidth]{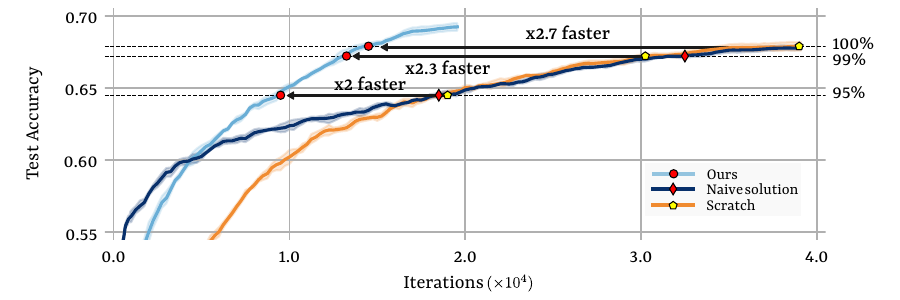}
	\caption{Test accuracy on CIFAR100 (70+30) with a model pre-trained on 70 classes. The `scratch' method starts from random initialization, while the `naive' approach uses the pre-trained model without modification. `Ours' modifies the optimization process (see Section~\ref{sec:method}) and matches the scratch performance with $2.7\times$ lower computational cost.}
	\label{fig:eye_catcher}
 \vspace{-0.4cm}
\end{figure}

Instead of the wasteful retraining from scratch approach, we propose to focus on continuous solutions, which leverage the existing model when new data becomes available. The simplest of these solutions is to continue the training of the available model on a combined dataset of old and new data. Even though this solution uses a pre-trained initialization, it converges at a similar speed as retraining from scratch (Figure~\ref{fig:eye_catcher}). Besides slower convergence, often worse performance is obtained when starting from a previous solution \citep{ash2020warm}, making the challenge of finding computational gains even more difficult. There are good reasons to study these problems in memory-restricted settings, without access to all old data. Yet as a first attempt to reduce the computational cost in continuous settings, we allow to store all old data in this paper. We believe by first tackling this problem, future work can solve the same problem with tighter restrictions on the memory.

In Section~\ref{sec:method}, we start from the canonical SGD update rule and show that all of its aspects – the initialization, the objective function, the sampling process, and the hyper-parameters – can significantly reduce computational costs.  We outline and evaluate several strategies -- inspired by the literature --  that act on these aspects. Each of them presents a promising avenue for future research to accelerate the convergence of continuous models. In Section~\ref{sec:results}, we show that while these methods individually reduce computational costs, their effects are to some extent complementary; combining them yields even greater reductions. In many of the tested scenarios, our approach leads to more than a two-fold decrease in computational complexity, compared to retraining from scratch. Such savings can make a substantial difference, especially when retraining happens repeatedly. Finally, we demonstrate that the proposed methods improve training efficiency across a variety of image classification datasets, multi-task settings, and domain incremental scenarios, highlighting the robustness and potential of our method to reduce the computational burden of re-training machine learning models.

\subsection{Our contribution}
\begin{itemize}
    \item We propose a novel way of evaluating continuous training, allowing models full access to previous data and measuring their computational cost rather than only their accuracy. 
    
	\item We discuss the different areas in the optimization process that can be explored to accelerate convergence. Each area is broad enough to be the focus of a different method.

	\item For each area, we introduce a first-step method based on the literature, which already significantly enhances the learning speed of models that are trained continuously.

	\item We demonstrate that improvements in these areas are complementary and applicable in many scenarios, meaning advancements in one area do not preclude further gains in others.
\end{itemize}

\subsection{Related Work}

\paragraph{Warm start.}
Starting from non-random initialization is most commonly used in transfer learning, where a pre-trained model (typically on ImageNet) serves as a starting point to kick-start training downstream tasks~\citep{zhuang2020comprehensive}. Contrary to our work, these works are often not concerned with performance on the pre-training (\ie old) data. While beneficial, pre-training may hurt downstream performance~\citep{zoph2020rethinking}. This may be explained by a loss of plasticity in trained networks~\citep{dohare2024loss, abbas2023loss}. \citet{ash2020warm} showed that when continuously training on the same data source, lower performance is reached than when starting from scratch. \citet{gupta2023continual, parmar2024reuse} study how to continually pre-train NLP models which is strongly related to our setting. They examine learning rate scheduling but do not consider loss of plasticity, regularization, and data selection as done in this paper.

\paragraph{Faster optimization.} At its core, machine learning involves solving a challenging and computationally expensive optimization problem, and many approaches have been proposed to streamline this process~\citep{sun2019survey}. First-order methods, such as stochastic gradient descent (SGD)\citep{robbins1951stochastic}, are the most widely used, where the full gradient is typically approximated using small batches. However, SGD can suffer from slow convergence, especially in high-variance settings\citep{johnson2013accelerating}, which can be mitigated by techniques like Nesterov momentum~\citep{sutskever2013importance}.  Adaptive approaches such as AdaGrad~\citep{duchi2011adaptive} and Adam~\citep{kingma2014adam} help by adjusting the learning rate dynamically, though explicit learning rate scheduling often enhances their performance~\citep{loshchilov2017sgdr, smith2019super}. Second-order methods, despite their promise, are hampered by the computational expense of estimating the Hessian~\citep{martens2016second}. \citet{goyal2017accurate} also highlights the intricate relationship between batch size and learning rate in neural network optimization. Additionally, regularization techniques like batch normalization~\citep{ioffe20215batch} and weight normalization~\citep{salimans2016weight} can further improve convergence. These considerations become especially important when optimizing from a pre-trained model, as is the case in this work, rather than from random initialization~\citep{narkhede2022review}.

\paragraph{Continual learning.} Continual learning concerns cases where data is not available all at once (see \eg~\citep{de2021continual,wang2024comprehensive} for surveys). Most studies have focused on settings with strong constraints on the amount of old data that can be stored~\citep{verwimpcontinual}. Replay methods~\citep{chaudhry2019continual, buzzega2020dark} aim to use this stored data as effectively as possible. However, even with replay mechanisms, learning new data often leads to (catastrophic) forgetting of previously learned examples. Other approaches modify the model architectures~\citep[\eg][]{yan2021dynamically} and regularization losses~\citep[\eg][]{li2017learning} to reduce forgetting. In contrast, some works have looked at settings where computational cost is restricted rather than memory costs. In these cases, standard replay outperforms other methods~\citep{prabhu2023computationally}. Later works~\citep{smith2024adaptive,harun2023grasp} improved replay techniques in these settings. Our work, however, imposes no memory restrictions and instead focuses on accelerating learning compared to models trained from scratch.

\section{Problem Description}

We consider the following setup: an existing dataset, denoted as $\mathcal{D}_{old}=(\mathcal{X}_{old}, \mathcal{Y}_{old})$, where $\mathcal{X}_{old}$ represents the input examples and $\mathcal{Y}_{old}$ the corresponding labels. A model $f_{old}:\mathcal{X}_{old}\rightarrow\mathcal{Y}_{old}$, has already been trained on this dataset. We are then provided with some new data, $\mathcal{D}_{new}=(\mathcal{X}_{new}, \mathcal{Y}_{new})$. The objective is to get a model $f:(\mathcal{X}_{old}\cup\mathcal{X}_{new})\rightarrow(\mathcal{Y}_{old}\cup\mathcal{Y}_{new})$ that performs well on both the new and the old datasets, $\mathcal{D}=\mathcal{D}_{old}\cup\mathcal{D}_{new}$, with a computational cost as low as possible.

To ensure a fair comparison between models, we use the same architecture when comparing the computation costs of different training methods. This, and keeping a fixed batch size, leads to a fixed number of FLOPs per iteration. Their equivalence allows us to report training iterations, which is easier to measure. Let $f^i$ represent the model $f$ after $i$ training iterations. We define the speed $s$ of a model $f$ in achieving a target accuracy, $a$, as the first iteration in which $f$ reached or surpassed that accuracy. Formally:

\begin{equation}
	s(f, a) = \min_i{\left\{i\in\mathbb{N}: \frac{1}{|\mathcal{D}|}\sum_j^{|\mathcal{D}|}\mathbbm{1}[f^i(x_j)=y_j]\geq a \right\}}
\end{equation}

We use the relative speed-up $L_r$ when comparing the performance of different models against a baseline model. The baseline model $f_{scratch}$ is trained from scratch on the combined dataset $\mathcal{D}$ and achieves an accuracy of $a_{scratch}$. Then we can define $L_r$ as:

\begin{equation}
	L_r(f, f_{scratch}) = \frac{s(f_{scratch}, a_{scratch})}{s(f, r / 100 \cdot a_{scratch})}
\end{equation}

or the relative number of iterations that a model $f$ requires to reach the same ($L_{100}$) or a fraction of (\eg $L_{99}$) the final accuracy of a model trained from scratch. For instance, $L_{99} = 2$ would indicate that model $f$ attains an accuracy of $0.99 \cdot a_{scratch}$ with a $2$ times lower computational cost than was required to train the full model from scratch to attain $a_{scratch}$, or equivalently, half the number of iterations. To reduce notation complexity, we will simply use $L_r$ in the remainder of the paper when the models can be inferred from context. Note that this measure only works when each iteration has the same computational cost, but the idea can easily be extended to when this is not the case.

\subsection{Implementation details}
\paragraph{Datasets.} We conducted experiments on a variety of image classification datasets, including CIFAR-10 \citep{krizhevsky2009learning}, CIFAR-100, subsets of ImageNet (ImageNet-100 and ImageNet-200) \citep{deng2009imagenet}, and Adaptiope \citep{ringwald2021adaptiope}. For continuous training, each dataset was divided into disjoint subsets, and training proceeded in a cumulative manner. For example, in CIFAR-100 (80+10+10), classes are split into three groups: 80 classes, followed by 10, and 10. The model was trained sequentially, first on the 80 classes, then on the combined 80 + 10, and finally on all 100 classes. Class splits were randomized, where the specific seeds are available in the attached code.

\paragraph{Training and baselines.} Unless otherwise specified, we used ResNet-18 with a cosine annealing scheduler \citep{loshchilov2017sgdr}, the Adam optimizer, and a learning rate of 0.001. All models are trained with standard cropping and horizontal flipping augmentations. Additional model and hyperparameter details can be found in the attached code and Appendix~\ref{sec:hyperparams}. The \textit{scratch} baseline indicates a model that is trained from a random initialization on both old and new data together. The \textit{naive} baseline represents a continuous model that simply continues training from the old model, without any modification.

\paragraph{Experimental details.} Every experiment shown in this paper is repeated five times. The results shown are the averages of these experiments, accompanied by their standard error in plots. The results presented are always obtained by using the combined test sets of the old and the new datasets.

\section{Method}
\label{sec:method}

In deep learning, optimization is typically performed using variants of stochastic gradient descent, where model parameters $\theta_i$ are updated by subtracting an estimate of the gradient of the objective function, based on a small batch of samples. The standard minibatch SGD update rule is:

\begin{equation}
	\theta_{i+1} = \theta_i - \frac{\eta}{N} \sum_{j\in B_i} \nabla L(\theta_i, (x_j, y_j))  
\end{equation}

where $\eta$ is the learning rate, $L$ is the objective function, and $B_i$ a batch of $N$ examples sampled from the full dataset $\mathcal{D}$. All components of this update -- model initialization ($\theta_0$), batch composition, learning rate adjustments, and modifications to the objective function -- play a critical role in the optimization trajectory and convergence speed. For other optimizers like \eg Adam~\citep{kingma2014adam}, the simple average here would be replaced by a momentum-based average, but the idea remains the same. In the following sections, we explore how each of these elements can significantly accelerate continuous model training. CIFAR-100 (70+30) serves as a case study for comparing various strategies.

\subsection{Initialization}
\label{sec:method:init}
\begin{figure}[tb]
	\centering
	\begin{minipage}[t]{.49\textwidth}
  	\centering
  	\includegraphics[width=1.0\linewidth]{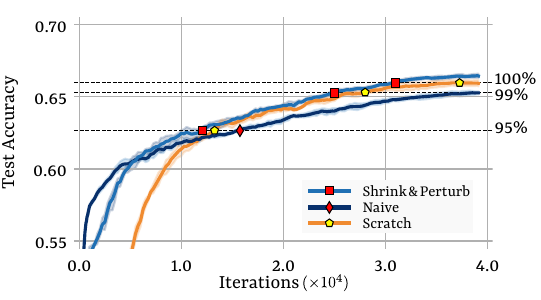}
  	\caption{\textbf{Initialization}. Naive continuous training is slower and less accurate than retraining from scratch. Re-introducing plasticity with shrink-and-perturb improves both speed and accuracy, surpassing scratch training.}
  	\label{fig:initialization}
	\end{minipage}%
	\hfill
	\begin{minipage}[t]{.49\textwidth}
  	\centering
  	\includegraphics[width=1.0\linewidth]{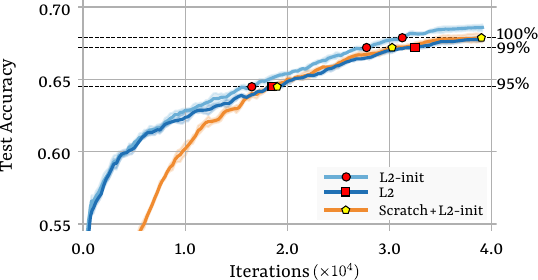}
  	\caption{\textbf{Objective function}. Regularizing the objective function with $L2$-losses is beneficial in both from scratch and continuous learning, yet the latter outperforms the former when using $L2$-init regularization.}
  	\label{fig:regularization}
	\end{minipage}
\end{figure}

When training models from scratch, careful random initialization offers several advantageous properties \citep{narkhede2022review}. However, during continuous training, many of these advantages are lost as training does not start from a random set of weights. Several works have shown that this potentially leads to reduced plasticity, \ie not being able to learn new information as fast and accurately as a model trained from scratch~\citep{ash2020warm, dohare2024loss}. This issue is similarly observed in continuous training, as shown in Figure~\ref{fig:initialization}. The naive benchmark is both worse and converges slower than retraining from scratch.

To restore plasticity during warm-start training without distribution shifts, \citet{ash2020warm} introduced the `shrink and perturb' method. Rather than using the previously trained model's weights directly, the weights are shrunk by a factor $\alpha$ and combined with a small portion of randomly initialized parameters $\theta_{random}$. The resulting initialization $\theta_{init}$ is computed as:

\begin{equation}
	\theta_{init} = \alpha \theta_{old} + \beta \theta_{random}
\end{equation}

In our experiments, we used $\alpha=0.4$ and $\beta=0.001$ without tuning, as proposed by the original authors. However, a broad range of $\alpha$ and $\beta$ values yielded qualitatively similar results (see Appendix~\ref{sec:init_abl}). In Figure~\ref{fig:initialization}, we compare ResNet-18 models trained continuously on CIFAR-100 (70+30) with and without the shrink-and-perturb method. The results show that re-introducing plasticity can not only accelerate convergence but also potentially improve final accuracy compared to both scratch training and continuous training without it.

\subsection{Objective function and regularization}

An alternative approach to tackle the reduced plasticity problem is to prevent it from arising in the first place by changing the objective function $L$. Inspired by the continual backpropagation idea of ~\citet{dohare2024loss}, \citet{kumar2023maintaining} propose a simplified version that uses an $L2$ regularizer towards the initial random weights $\theta_0$, rather than towards the origin as is typically done in $L2$ regularization. In our setting, this involves modifying the objective function $L$ to include this $L2$-init regularizer:

\begin{equation}
	L_{reg}(\theta_i, (x_j, y_j)) =  L(\theta_i, (x_j, y_j)) + \lambda \|\theta_i - \theta_0 \|^2
\end{equation}

In our experiments, we used $\lambda=0.01$ without tuning, as proposed by the original authors. However, a broad range of $\lambda$ values yielded qualitatively similar results (see Appendix~\ref{sec:reg_abl}). In Figure~\ref{fig:regularization}, we train ResNet-18 models on CIFAR-100 (70+30) and compare the results of models trained from scratch with those trained continuously, both with the $L2$-init regularizer. The results indicate that adding this regularization term accelerates the convergence of the continuously trained models, more so than it improves models trained from scratch. Since regularization can also benefit models trained from scratch, we use this baseline here, as well as the more standard $L2$ regularization (\ie $\theta_0 = \textbf{0}$) which is not as effective as $L2$-init regularization.

\subsection{Batch composition}

\begin{figure}[tb]
	\centering
	\begin{minipage}[t]{.49\textwidth}
  	\centering
  	\includegraphics[width=1.0\linewidth]{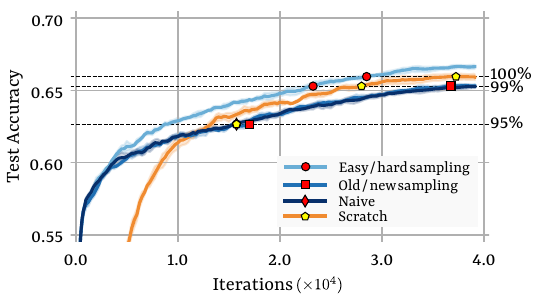}
  	\caption{
   \textbf{Batch composition}. `Old / new' sampling balances old and new examples in each batch, unlike the naive baseline, which uses proportional sampling. `Easy / hard' sampling reduces the inclusion of the easiest and hardest old examples, significantly improving performance. (Naive and 'old/new' results nearly overlap.)}
  	\label{fig:batch_composition}
	\end{minipage}%
	\hfill
	\begin{minipage}[t]{.49\textwidth}
  	\centering
  	\includegraphics[width=1.0\linewidth]{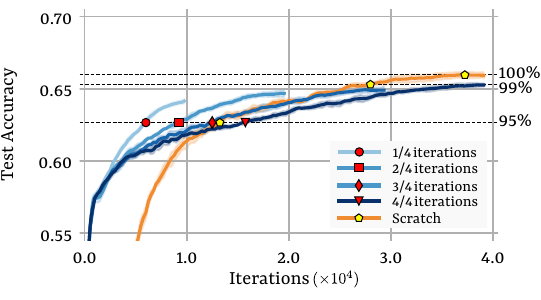}
  	\caption{\textbf{Hyperparameters}. Shortening the learning rate scheduler allows for faster convergence but at the cost of lower final accuracy. On its own, changing the scheduler does not reach the required accuracy, yet when combined with the other aspects, it becomes important (see Section~\ref{sec:ablation})}
  	\label{fig:hyperparameters}
	\end{minipage}
\vspace{-1em}
\end{figure}

Estimating the full gradient of a dataset is expensive in the deep learning case, as many iterations are required to reach a good solution. To overcome this, minibatches containing a small part of the entire dataset are used to estimate the gradient. Typically, examples are sampled from the full dataset with uniform probability. In continual learning methods, it is a common practice to balance examples from old and new data in a batch~\citep{rolnick2019experience}, which either increases or decreases the sampling probability of old data depending on whether there is either more or less old than new data.

In Figure~\ref{fig:batch_composition}, we show that having an equal number of new and old examples (`old / new sampling') does not improve the results compared to the naive approach, which balances old and new examples in a batch according to their ratio in the full dataset (\ie 70\% old examples and 30\% new ones in this example). \citet{katharopoulos2018not} show that the optimal sampling distribution is proportional to the gradient norms of the individual examples. In Appendix~\ref{sec:grad_norms}, we show that at the very start of training the gradient norms of the old examples are indeed smaller, but after about 100 iterations, there is no noticeable difference on the class level, which explains the results.

While the ratio of old and new examples does not have an immediate effect, not all data in the replay memory is equally useful. Often in continual learning, access to the old dataset $\mathcal{D}_{old}$ is limited to a small fraction. In our case, the replay memory has infinite capacity. We build on the work of~\citet{hacohen2024forgetting}, who propose a sampling strategy that reduces the importance of very easy and very difficult examples in the memory. More formally, they define learning speed $ls$ of a sample $x_j$ as the relative epoch in which a sample is classified correctly:

\begin{equation}
	ls(x_j, y_j) = \frac{1}{E} \sum_{i=1}^E \mathbbm{1}[f^i(x_j) = y_j]
\end{equation}

with $E$ the total number of epochs and $f^i$ the model at epoch $i$.

The learning speed is used to order the old examples from easy to hard, given that the necessary information is recorded during training of the old model (For more details, see Appendix~\ref{sec:batch_abl}). Using this order, we reduce the sampling probability of the $10\%$ easiest and highest examples to one-tenth of the other examples. In the easy examples, there is no information left, and the hardest ones are never learned, and thus as useful. The results in Figure~\ref{fig:batch_composition} show that this approach (`easy / hard sampling') is helpful and speeds up training in the continuous case. In Appendix~\ref{sec:batch_abl} we show that this is robust to the exact hyperparameters and that, while sacrificing some convergence speed, the easiest and hardest examples can be removed entirely.

\subsection{Learning rate scheduling}
The learning rate controls the step size in stochastic gradient descent, directly influencing convergence speed. In deep learning, the learning rate typically starts high to allow faster progress toward optimal solutions and decreases gradually to avoid overshooting~\citep{zeiler2012adadelta}. Common scheduling strategies include `multistep' scheduling~\citep{zagoruyko2016wide}, which reduces the learning rate at set intervals, and cosine annealing~\citep{loshchilov2017sgdr}, which decays it more smoothly over time.

When training a model continuously, since the model is already trained on parts of the data, a more aggressive learning rate scheduling can accelerate convergence. While the learning rate still needs to decrease over time, this can happen more rapidly over fewer iterations, as the model has already learned key information. As shown in Figure~\ref{fig:loss_plateau} in the Appendix, the loss curve for the continuous model reaches a plateau much faster compared to scratch training, indicating that the model approaches a local optimum more quickly. This supports the need for a faster reduction in the learning rate~\citep{zeiler2012adadelta}.

We train ResNet-18 models on CIFAR-100 (70+30 split), using different lengths of cosine learning rate schedulers. The most aggressive scheduler used only $25\%$ of the total iterations, while others used $50\%$ and $75\%$. The learning curves, depicting the mean test errors from these experiments, are shown in Figure~\ref{fig:hyperparameters}. The results indicate that more aggressive schedulers lead to faster convergence of the continuous model to a performance level comparable to the scratch solution. However, overly aggressive scheduling hurts the final performance, as the networks fail to achieve $100\%$ of the scratch model’s final accuracy. On its own, reducing the learning length is not helpful, yet when combined with the other aspects, it becomes important (See Section~\ref{sec:ablation}). Similar qualitative results were observed with multistep scheduling, see Appendix~\ref{sec:sched_abl}. For the remainder of this paper, any adjustments to the learning rate scheduler for specific methods are explicitly mentioned, and a comparison to the unmodified variant is provided.

\section{Results}
\label{sec:results}

In the previous section, we explored several key aspects of SGD optimization, and each can serve as the foundation for methods that reduce the computational cost of continuous training. By leveraging insights from various works in the literature, we proposed a method for each aspect, demonstrating the effectiveness of each method individually.

In this section, we begin by showing that these methods are complementary, with their combination further accelerating convergence beyond what is achieved by applying them separately. We show that although the same techniques can lead to better convergence when training from scratch, the benefit is larger in the continuous setting. We then extend this combined approach to various datasets, scenarios, and data splits, demonstrating its applicability across image classification tasks.

\begin{table}[tb]
\caption{Ablation results of the four different aspects studied in Section~\ref{sec:method} under `\textit{continuous}' on CIFAR100 (70+30). All speed-ups are relative to the model trained from scratch without any modification,  `/' indicates the required accuracy was not reached. Each of the aspects individually establishes a speed-up, and combining them improves the results further, albeit not fully cumulative. On the right hand side of the table, `\textit{scratch}' shows the result of a model trained from scratch using the same techniques. Some of them improve the training speed, but to a lesser extent than in the continuous case.}
\label{tab:ablation}
\resizebox{\columnwidth}{!}{%
\begin{tabular}{@{}cccc|ccc|lll@{}}
 &  &  &  & \multicolumn{3}{c|}{\textbf{Continuous}} & \multicolumn{3}{c}{\textbf{Scratch}} \\
\textbf{Initialization} & \textbf{Regularization} & \textbf{Data} & \textbf{Scheduler} & \textbf{Max Acc} & $L_{99}$ & $L_{100}$ & \textbf{Max Acc} & $L_{99}$ & $L_{100}$ \\ \midrule
 &  &  &  & $65.26$ & / & / & $65.92$ & $\times 1.33$ & $\times 1.00$ \\ \midrule
\checkmark &  &  &  & $66.43$ & $\times 1.49$ & $\times 1.20$ & $65.06$ & / & / \\
 & \checkmark &  &  & $68.60$ & $\times 1.94$ & $\times 1.66$ & $67.90$ & $\times 1.69$ & $\times 1.51$ \\
 &  & \checkmark &  & $66.65$ & $\times 1.60$ & $\times 1.31$ & $66.61$ & $\times 1.54$ & $\times 1.31$ \\
 &  &  & $\times 0.25$ & $64.15$ & / & / & $63.72$ & / & / \\ \midrule
\checkmark & \checkmark &  &  & $68.91$ & $\times 2.19$ & $\times 1.82$ & $68.01$ & $\times 1.75$ & $\times 1.54$ \\
\checkmark &  & \checkmark &  & $66.76$ & $\times 2.92$ & $\times 2.53$ & $66.61$ & $\times 1.54$ & $\times 1.31$ \\
 & \checkmark & \checkmark &  & $68.86$ & $\times 1.96$ & $\times 1.69$ & $68.30$ & $\times 1.67$ & $\times 1.49$ \\
\checkmark & \checkmark & \checkmark &  & $\mathbf{69.47}$ & $\times 2.19$ & $\times 1.84$ & $68.31$ & $\times 1.69$ & $\times 1.52$ \\
\checkmark & \checkmark & \checkmark & $\times 0.25$ & $68.26$ & $\mathbf{\times 5.73}$ & $\mathbf{\times 5.32}$ & $63.74$ & / & / \\ \bottomrule
\end{tabular}%
}
\end{table}

\subsection{Ablations}
\label{sec:ablation}
In the previous section, we introduced several methods aimed at speeding up the convergence of continuous models, each targeting a different aspect of the optimization process. We now examine whether these improvements are complementary or if the benefits of one method negate the effectiveness of others. Table~\ref{tab:ablation} presents results from training five ResNet-18 models and compares all combinations. While not fully cumulative, each combination offers benefits, either in convergence speed or in final accuracy. Additionally, we applied the same techniques to a model trained from scratch, to rule out the possibility that the techniques are \emph{only} improving overall learning capabilities. 

The results show that while some optimization aspects have an effect on the scratch models as well (most notably regularization), their influence is always stronger in the continuous case. This is especially clear when shortening the learning rate scheduler, which allows the continuous model to learn faster, yet greatly reduces the final accuracy of the from-scratch model. These results indicate that the proposed techniques are effective in leveraging the benefits of having a model trained on the old data available. Future research targeting different aspects of the continuous learning process could yield further gains in training speed, as multiple optimization strategies can be effectively integrated to speed up the learning.

\subsection{Multiple tasks}
\begin{figure}[tbh]
	\centering
	\begin{minipage}[t]{.49\textwidth}
  	\centering
  	\includegraphics[width=1.0\linewidth]{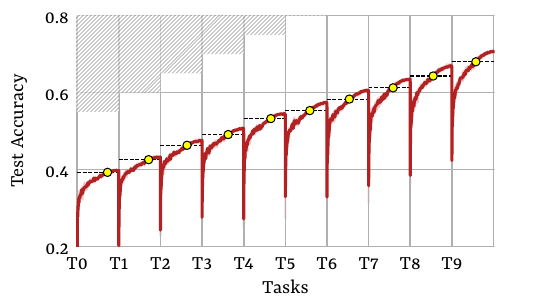}
  	\caption{Test accuracy on the full CIFAR-100 when training a continuous model with our method on CIFAR100 $(50+\sum_{i=1}^{10}5)$. Yellow dots mark the $L_{100}$ iteration, where the continuous model outperforms the from-scratch model. The shaded area indicates maximum possible performance based on data availability (e.g., $55\%$ during the first task $50+5$).}
  	\label{fig:multi-task}
	\end{minipage}%
	\hfill
	\begin{minipage}[t]{.49\textwidth}
  	\centering
  	\includegraphics[width=1.0\linewidth]{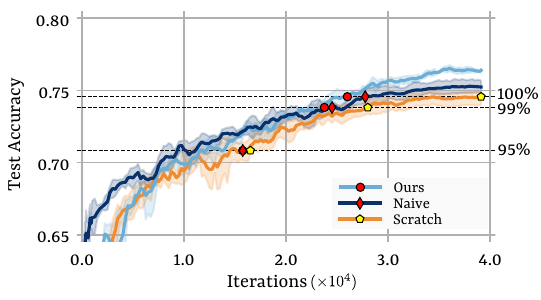}
  	\caption{Domain adaptation results using the Adaptiope dataset. The `old' data contains \textit{product images}, the `new' data has \textit{real life} images. The test accuracy is on both domains. Although naively continuing to train is nearly as fast as our method in this setting, our final test accuracy is considerably higher.}
  	\label{fig:domain_adaptation}
	\end{minipage}
\end{figure}

We used CIFAR100 (70+30) as a case study in Section~\ref{sec:method}, adding 30 new classes. In many continual learning scenarios, new data is not only added once, but repeatedly. In Figure~\ref{fig:multi-task} we show that applying the same aspects on consecutive tasks continues to work as in the one-task case. For each task, we also train a model from scratch on all classes available up to this point. In Figure~\ref{fig:multi-task}, the yellow dots indicate when the continuous model reaches the same performance ($L_{100}$) as the from-scratch model that is trained on that task and all previous data. For every task $L_{100}$ is larger than one and gets progressively larger. This indicates that our methods is repeatedly applicable, and benefits from larger total accumulated learning time.

These results highlight an important insight: in total, the continuous models have trained for more iterations than the models that are trained from scratch, which also explains why their final accuracy may surpass that of training from scratch. However, when accumulating past costs, the total cost of the continuous model is significantly lower than the sum of costs for models trained from scratch for every task.

\subsection{Domain adaptation}
All previous experiments had new classes in the `new' data, often referred to as class-incremental learning in continual learning~\citep{de2021continual}. Here, we use the Adaptiope dataset~\citep{ringwald2021adaptiope} to show that our method also works when new data contains no new classes, but the same classes from a different domain. In particular, the `old' data consists of $123$ categories of product images from a shopping website and the `new' data are real-life images of the same products captured by users. The objective is to achieve strong performance across both domains, ideally with faster convergence than re-training the model from scratch on both datasets.

Figure~\ref{fig:domain_adaptation} shows how our method outperforms both training from scratch and the naive continuous approach when including a new domain. The relative speed-up compared to the naive baselines is smaller than in the class-incremental case, but the final accuracy is considerably improved.

\begin{table}[htb]
	\centering
	\caption{
	Results of our method across various datasets and splits. We report final test accuracy and relative speedup to reach similar performance compared to the scratch solution. The multiplier after the algorithm name indicates learning rate scheduling aggressiveness (e.g., Ours ($\times 0.5$) means the minimum learning rate is achieved at half the iterations of Ours ($\times 1$)). `/' indicates that the accuracy level was not reached. Table (a) presents results on different datasets, while Table (b) provides an in-depth analysis of CIFAR-100 with various class splits compared to the better scratch + L2-init model, hence the slightly lower results compared to Table~\ref{tab:ablation}}
	\begin{subtable}[t]{.48\linewidth}
    	\centering
    	\caption{More image classification benchmarks}
    	\label{tab:different_benchmarks}
    	\resizebox{\columnwidth}{!}{%
    	\begin{tabular}{@{}llccc@{}}
    	\toprule
    	\textbf{Dataset}	& \textbf{Algorithm}	& \textbf{Max Acc} & $L_{99}$ & $L_{100}$ \\
    	\midrule
    	\multirow{4}{*}{\makecell{CIFAR10 \\ (8+2)}}
    	& Scratch          	& 83.32 & $\times 1.55$ & $\times 1$  \\
    	& Continuous       	& 81.37 & $/$ & $/$  \\
    	& Ours ($\times 1$)	& 83.95 & $\times 1.83$ & $\times 1.29$  \\
    	& Ours ($\times 0.5$)  & \textbf{84.33} & $\times 3.61$ & $\times 2.92$ \\
    	\midrule
    	\multirow{4}{*}{\makecell{Adaptiope-PI \\ (100+23)}}
    	& Scratch          	& 74.59 & $\times 1.14$ & $\times 1$  \\
    	& Continuous       	& 74.93 & $\times 1.17$ & $1.02$  \\
    	& Ours ($\times 1$)	& \textbf{77.51} & $\times 1.55$ & $\times 1.42$  \\
    	& Ours ($\times 0.5$)  & 76.75 & $\times 2.36$ & $\times 2.24$  \\
    	\midrule
    	\multirow{4}{*}{\makecell{ImageNet-100 \\ (80+20)}}
    	& Scratch          	& 62.34 & $\times 1.18$ & $\times 1$  \\
    	& Continuous       	& 62.21 & $\times 1.23$ & $/$  \\
    	& Ours ($\times 1$)	& \textbf{67.41} & $\times 2.16$ & $\times 1.97$  \\
    	& Ours ($\times 0.5$)  & 66.26 & $\times 3.39$ & $\times 3.09$  \\
    	\midrule
    	\multirow{4}{*}{\makecell{ImageNet-200 \\ (180+20)}}
    	& Scratch          	& 55.16 & $\times 1.18$ & $\times 1$\\
    	& Continuous       	& 53.5 & $/$ & $/$  \\
    	& Ours ($\times 1$)	& \textbf{59.62} & $\times 1.69$ & $\times 1.64$  \\
    	& Ours ($\times 0.5$)  & 58.54 & $\times 2.84$ & $\times 2.75$  \\
    	\bottomrule
    	\end{tabular}%
    	}
	\end{subtable}
	\hfill
	\begin{subtable}[t]{.495\linewidth}
	\caption{Different ratios of old and new classes.}
	\label{tab:different_splits}
	\resizebox{\columnwidth}{!}{%
    	\begin{tabular}{@{}llccc@{}}
    	\toprule
    	\textbf{Dataset}      	& \textbf{Algorithm} 	& \textbf{Max Acc} & $L_{99}$  	& $L_{100}$ 	\\ \midrule
    	\multirow{2}{*}{CIFAR100} & Scratch + L2-init  	& $67.90$      	& $\times 1.29$ & $\times 1.00$ \\
                              	& Scratch ($\times 0.5$) & $67.14$      	& /         	& /         	\\ \midrule
    	\multirow{3}{*}{90+10}	& Ours ($\times 1.0$) 	& $69.42$      	& $\times 1.56$ & $\times 1.41$ \\
                              	& Ours ($\times 0.5$) 	& $69.07$      	& $\times 2.95$ & $\times 2.65$ \\
                              	& Ours ($\times 0.25$)	& $68.42$      	& $\times 5.05$ & $\times 4.47$ \\ \midrule
    	\multirow{3}{*}{70+30}	& Ours ($\times 1.0$) 	& $69.47$      	& $\times 1.61$ & $\times 1.45$ \\
                              	& Ours ($\times 0.5$) 	& $68.77$      	& $\times 2.84$ & $\times 2.56$ \\
                              	& Ours ($\times 0.25$)	& $68.26$      	& $\times 4.74$ & $\times 4.34$ \\ \midrule
    	\multirow{3}{*}{50+50}	& Ours ($\times 1.0$) 	& $69.09$      	& $\times 1.50$ & $\times 1.35$ \\
                              	& Ours ($\times 0.5$) 	& $68.42$      	& $\times 2.70$ & $\times 2.37$ \\
                              	& Ours ($\times 0.25$)	& $67.57$      	& $\times 4.34$ & /         	\\ \midrule
    	\multirow{3}{*}{30+70}	& Ours ($\times 1.0$) 	& $68.81$      	& $\times 1.49$ & $\times 1.31$ \\
                              	& Ours ($\times 0.5$) 	& $68.10$      	& $\times 2.56$ & $\times 2.30$ \\
                              	& Ours ($\times 0.25$)	& $66.93$      	& /         	& /         	\\ \bottomrule
    	\end{tabular}%
	}
	\end{subtable}%
\vspace{-1em}
\end{table}

\subsection{Other datasets and scenarios}
In Section~\ref{sec:method}, we demonstrated the effectiveness of each method using the CIFAR-100 (70+30) dataset for consistency across experiments. Table~\ref{tab:different_benchmarks} extends these results to a range of datasets, including CIFAR-10, ImageNet-100, ImageNet-200, and the product images of Adaptiope. For each dataset, we compare the performance of models trained from scratch, a naive continuous model, and a continuous model using our full method. We report the relative speed-up to reach $99\%$ and $100\%$ of the scratch-trained model’s accuracy, along with the final accuracy. Across all the different datasets we tested, our method consistently enhances the speed of convergence both for the $L_{99}$ and $L_{100}$ cases, often also surpassing the final accuracy of the scratch model.

Table~\ref{tab:different_splits} presents results for different class split ratios, including CIFAR-100 splits of (90+10), (70+30), (50+50), and (30+70). Our methods consistently accelerate training, with speed-up being more significant when the second data split is smaller. This aligns with intuition: when the initial `old' data is larger, the old model already has more knowledge, requiring fewer updates when exposed to new data, compared to training from scratch, which treats both splits equally.

\section{Conclusion}
In this work, we have shown that the computational cost that is paid to train models on old data should not be treated as something that is lost, but rather as a starting point for further training whenever new data becomes available. We used all old data that was available. Future work can either focus on removing that constraint step-by-step, or further improving the computational gains we started. Throughout this paper, we studied the four main components of the traditional SGD update rule -- the initialization, objective function, data selection and hyperparameters -- and showed that each of them individually can contribute to faster convergence on old and new data combined when starting from an old model. These aspects are thoroughly tested to be robust and effective across a wide range of scenarios. Our proposals should be seen as starting points to work further towards solutions that are even more effective in re-using old models, and saving computational costs across the board in machine learning development and applications. 


\bibliography{main}
\bibliographystyle{iclr2025_conference}

\appendix
\section{Appendix}
\subsection{Cost estimation}
\label{sec:cost_estimation}
To train a machine learning model there is both a memory (storage) and compute cost. Here we will work out an example for training a Vision Transformer (ViT)~\citep{vaswani2017attention} on ImageNet 21k. This dataset is about $1.13$ terabytes. Cloud storage for common providers averages around $0.023\$$ per GB each month, which would be around 30.13\$ per month~\citep{googlePricingCloud}. However, local storage is much cheaper than cloud storage. Today hard disk storage is about $11$\$ per TB~\citep{ourworldindataStorage}, which can last for 10 years or more.

ViT models were trained on 8 Nividia P100 gpus for 3.5 days~\citep{vaswani2017attention}, which cost \$1.46 per hour on Google Cloud, which would be \$981 in total for the entire training~\citep{googlePricingCompute}. Buying the GPUs would be much more expensive, with prices for a single A100 GPU above \$$10.000$~\citep{nvidiaNVIDIAA100}. Modern A100 GPUs are about $3.5$ times faster on real-life work loads~\citep{nesiTechInsights}, but more expensive to rent. At 4\$$4.05$ per hour, the total price would still be \$777.5.

\subsection{Hyperparameters}
\label{sec:hyperparams}
Unless specified otherwise, all experiments are trained with the Adam optimizer~\citep{kingma2014adam}, with default settings of $\beta_1=0.9$, $\beta_2 = 0.999$ and no weight decay. The starting learning rate is equal to $0.001$ and the batch size is consistently $128$ in all experiments. Unless specified differently, a cosine scheduler is used where the minimal learning rate $1e-6$ is reached after $39.100$ iterations, which is equal to $50$ epochs using the complete CIFAR100 dataset. All experiments use cropping and random horizontal flip augmentations.

\subsection{Initialization}
\label{sec:init_abl}
\begin{figure}[H]
	\centering
	\begin{minipage}[t]{.49\textwidth}
  	\centering
  	\includegraphics[width=1.0\linewidth]{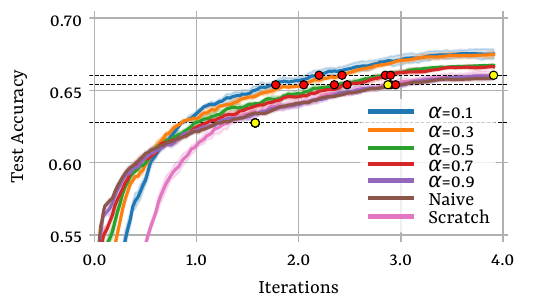}
  	\caption{Grid test of the $\alpha$ (shrink) hyperparameter in the shrink and perturb update rule. }
  	\label{fig:snp_abl}
	\end{minipage}%
	\hfill
	\begin{minipage}[t]{.49\textwidth}
  	\centering
  	\includegraphics[width=1.0\linewidth]{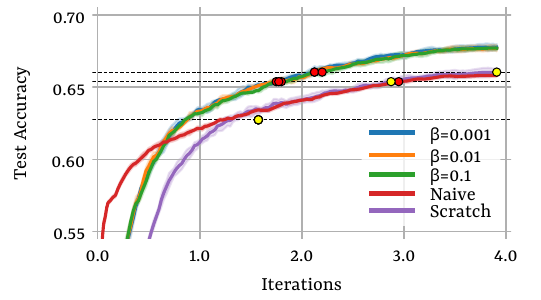}
  	\caption{Grid test of the $\beta$ (perturb) hyperparameter in the shrink and perturb update rule}
  	\label{fig:snp_shrink_abl}
	\end{minipage}
\end{figure}

Figure~\ref{fig:snp_abl} shows different values of $\alpha$ in the shrink and perturb update rule. In general, shrinking more gives better results, although it slows down learning results at the start of learning. Shrinking enough is necessary and not doing so may lead to suboptimal accuracies because of loss of plasticity. In Figure~\ref{fig:snp_shrink_abl} different values of $\beta$ are tested, which add various levels of noise to the initialization. There is no visible effect of the size of this parameter, which is roughly in line with~\citep{ash2020warm}.

\subsection{Regularization}
\label{sec:reg_abl}
\begin{figure}[H]
	\centering
	\includegraphics[width=0.65\linewidth]{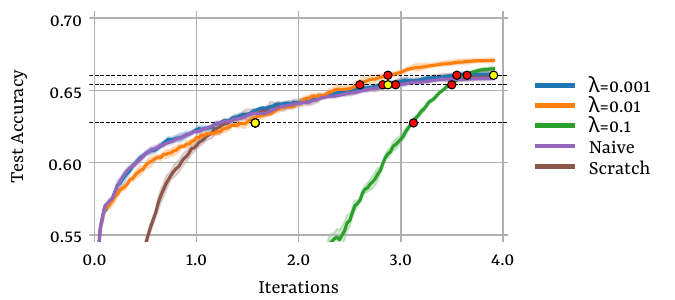}
	\caption{Ablation of the $\lambda$ parameter in the $L2$-init regularization.}
	\label{fig:l2_abl}
\end{figure}

Figure~\ref{fig:l2_abl} shows the influence of the $\lambda$ parameter in the $L2$-init regularization, which controls the strength of the regularization. A too low value will not have any effect, while setting this value too high, will lead to slower than necessary learning speeds.

\subsection{Batch composition}
\label{sec:batch_abl}
\begin{figure}[H]
\centering
\begin{minipage}[t]{.49\textwidth}
  \centering
  \includegraphics[width=1.0\linewidth]{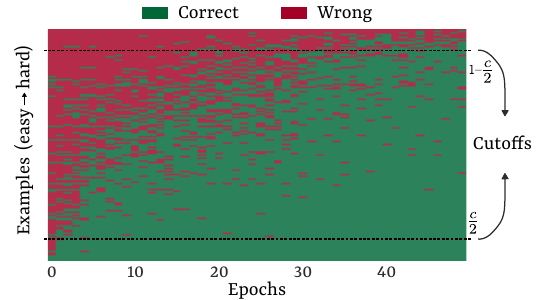}
  \caption{How fast examples are learned during the initial learning phase of training on the old data of CIFAR100 70+30 (i.e. the first 70 classes). Some examples are almost learned instantly (bottom), others never (top). The cutoffs indicate both the 10\% easiest and hardest examples.}
  \label{fig:learning_order}
\end{minipage}%
\hfill
\begin{minipage}[t]{.49\textwidth}
  \centering
  \includegraphics[width=1.0\linewidth]{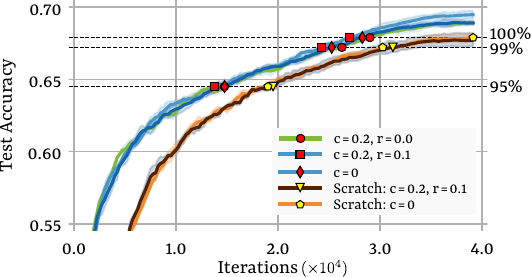}
  \caption{Test accuracy of CIFAR100 on the $70+30$ benchmark. Removing the easiest and hardest examples ($r=0$) barely has an influence. Sampling them with a low probability ($r=0.1$) allows to learn a little faster on the continuous model, without such an effect on the from scratch model.}
  \label{fig:sample_v2}
\end{minipage}
\end{figure}

Figure~\ref{fig:learning_order} shows how fast examples are learned during the first task of CIFAR100 (70+30). The x-axis shows the epochs, while each row indicates a single sample. The examples on the bottom are green almost from the start of training, these are the ones that are very easy: the model almost has no difficulty learning them, which also means they do not carry a lot of information. The ones on the top are almost completely red: they are never learned. These examples are equally not very useful: they are so hard the model can not learn them anyway (which may be a result of bad labeling, bad images etc.).

In Figure~\ref{fig:sample_v2}, we ablate two of the parameters of the `easy / hard' sampling process. $c$ indicates what proportion of the easy and hard examples is influenced. \eg $c=0.2$ means that 20\% of the examples is affected, or the 10\% easiest and the 10\% hardest. $r$ indicates the probability that the easy and hard examples are sampled in a batch, relative to the examples in the middle. \eg with $r=0.1$, an easy or hard example is 10 times less likely to be in a minibatch. The result with $r=0$ shows that we can even get rid of these examples, reducing the memory requirements for this method. When $r=0$, this becomes the method proposed by~\citep{hacohen2024forgetting}, on which this idea is based.

\subsection{Schedulers}
\label{sec:sched_abl}
\begin{figure}[H]
	\centering
	\includegraphics[width=0.5\linewidth]{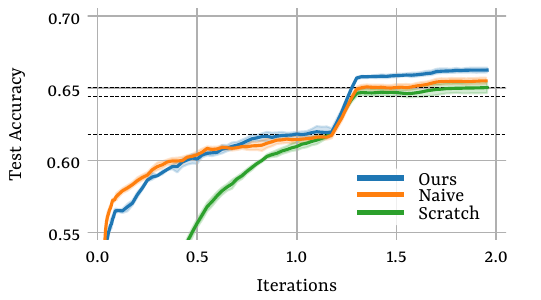}
	\caption{When using multi-step schedulers the qualitative result stays the same, although cosine schedulers are better suited to this task of continuous learning.}
	\label{fig:multistep}
\end{figure}

Figure~\ref{fig:multistep} shows an experiment on CIFAR100 (70+30), with a multistep learning rate scheduler (which reduces the learning rate by a fixed factor at fixed steps) rather than a cosine learning rate scheduler. This scheduler reaches the same conclusion, although in this case the iterations where the scheduler kicks in has a much larger influence than in the cosine scheduler case. In the continuous case, the learning rate is kept too high for too long, preventing the model to learn the last details, while the from scratch model is still learning.

\subsection{Gradient Norms}
\label{sec:grad_norms}
\begin{figure}[H]
	\centering
	\includegraphics[width=0.60\linewidth]{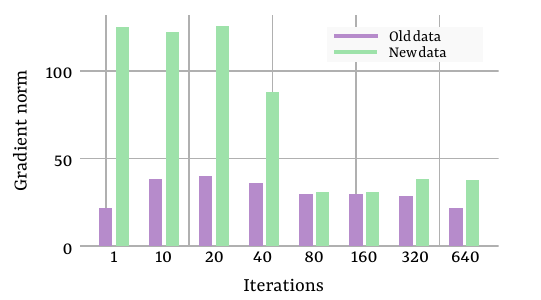}
	\caption{Gradient norms of `old' and `new' data during training of the naive baseline on CIFAR100 (70+30).}
	\label{fig:grad_norms}
\end{figure}

Figure~\ref{fig:grad_norms} shows the average gradient norm of examples of old and new data in the CIFAR100 (70+30) baseline during training of the naive baseline. While at the very start the gradient of new data is considerably higher, this difference has completely disappeared after more than 100 iterations, which is nearly immediately considering $40.000$ iterations in total. This explains mostly why it is not useful to oversample new data, which would indicate that new data is more important to learn than remembering the old data, which is not the goal.

\subsection{Loss plateau}
\label{sec:loss_plateau}
\begin{figure}[H]
	\centering
	\includegraphics[width=0.60\linewidth]{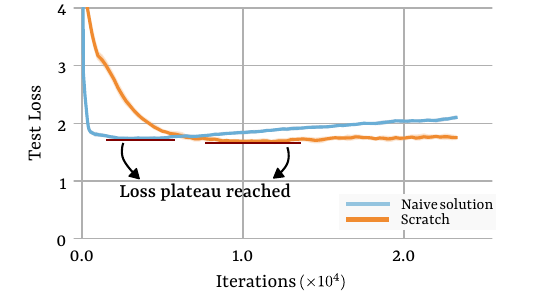}
	\caption{The continuous naive baseline loss plateaus earlier than that of the from scratch baseline.}
	\label{fig:loss_plateau}
\end{figure}

Figure~\ref{fig:loss_plateau} shows how the loss of the naive solution plateaus earlier than when training from scratch, thus allowing to schedule learning rates earlier. However, it is not because such a loss plateau is reached that the final result will be better, it merely indicates that the results won't get any better when training longer than from the point the plateau is reached. In fact, as can be seen in Figure~\ref{fig:loss_plateau}, the loss starts to increase again as overfitting takes place. In this sense, it might even be necessary to schedule early enough to obtain the best possible results.

\end{document}